\documentclass{article} 
\usepackage{iclr2020_conference,times}

\usepackage{amsmath,amsfonts,bm}

\def\eqref#1{equation~\ref{#1}}

\def\1{\bm{1}}

\def\rx{{\textnormal{x}}}

\DeclareMathAlphabet{\mathsfit}{\encodingdefault}{\sfdefault}{m}{sl}
\SetMathAlphabet{\mathsfit}{bold}{\encodingdefault}{\sfdefault}{bx}{n}

\newcommand{\E}{\mathbb{E}}

\usepackage{hyperref}
\usepackage{url}
\usepackage{array,multirow,graphicx}
\usepackage{booktabs}

\renewcommand{\headrulewidth}{0pt}

\usepackage{adjustbox}

\usepackage{enumitem}
\usepackage{wrapfig}
\usepackage{subcaption}

\title{Why Should we Combine Training and \\ Post-Training Methods for \\ Out-of-Distribution Detection?}

\iclrfinalcopy

\author{Aristotelis-Angelos Papadopoulos \& Nazim Shaikh \& Mohammad Reza Rajati \\
University of Southern California\\
Los Angeles, CA 90089, USA \\
\texttt{\{aristop,nshaikh,rajati\}@usc.edu} \\
}

\begin{document}

\maketitle
\begin{abstract}
Deep neural networks are known to achieve superior results in classification tasks. However, it has been recently shown that they are incapable to detect examples that are generated by a distribution which is different than the one they have been trained on since they are making overconfident prediction for Out-Of-Distribution (OOD) examples. OOD detection has attracted a lot of attention recently. In this paper, we review some of the most seminal recent algorithms in the OOD detection field, we divide those methods into training and post-training and we experimentally show how the combination of the former with the latter can achieve state-of-the-art results in the OOD detection task.\footnote{Preprint. Work in progress.}   
\end{abstract}

\section{Introduction}
Since the seminal work of \citet{NIPS2012_4824}, Deep Neural Networks (DNNs) have demonstrated great success in several applications, e.g. image classification, speech recognition, natural language processing etc. However, one of the most challenging tasks which has attracted a lot of attention recently is the Out-Of-Distribution (OOD) detection ability of a DNN, i.e. how to make a DNN able to detect examples that are generated by a probability distribution which is completely different from the one that has generated the examples that has been trained on.  

\citet{DBLP:conf/cvpr/NguyenYC15} showed that deep neural networks can make overconfident predictions for OOD examples while at the same time, overconfident predictions have been also related to overfitting problems \citep{DBLP:journals/corr/SzegedyVISW15}. The problem that DNNs are making predictions with a probability that is higher than their accuracy has been addressed to the literature as miscalibration \citep{Guo:2017:CMN:3305381.3305518}. These works have motivated many researchers in the field to develop algorithms to make a DNN capable of detecting OOD examples.

In this paper, we are interested about threshold-based OOD detection algorithms, i.e. algorithms that identify an example as in-distribution if the output probability of the DNN is above a threshold $\tau$ and as OOD if the output probability is below $\tau$. After classifying some of the most seminal recent threshold-based OOD detection algorithms based on whether they are training or post-training methods similar to \citet{anonymous2020zeroshot}, we combine the Outlier Exposure with Confidence Control (OECC) method proposed by \citet{anonymous2020simultaneous} which is a training method and to the best of our knowledge, is the state-of-the-art OOD detection algorithm for this category of methods, with the Mahalanobis distance-based classifier \citep{Lee:2018:SUF:3327757.3327819} and the zero-shot OOD detection method with feature correlations \citep{anonymous2020zeroshot} which both are post-training methods. We experimentally show that the combination of the aforementioned methods achieves state-of-the-art results in the OOD detection task demonstrating the potential of combining training and post-training methods for OOD detection in the future research efforts.


\section{Related Work}
\paragraph{Training Methods for OOD detection.} \citet{hendrycks17baseline} proposed a baseline for detecting misclassified and out-of-distibution examples based on their observation that the prediction probability of out-of-distribution examples tends to be lower than the prediction probability for correct examples. This was a premilinary work in the field of OOD detection and their method did not require access to the test distribution since they trained a neural network using only the cross-entropy loss function. \citet{2017arXiv171109325L} proposed to add a regularizer in the loss function while training the DNN. The regularizer used was the Kullback-Leibler (KL) divergence metric between the output distribution produced by the softmax layer of the DNN and the uniform distribution. The intuition behind the use of this loss function was to make the DNN predict in-distribution examples with high confidence while being highly uncertain about its predictions in OOD examples. In such a manner, a threshold $\tau$ could be successfully applied to distinguish in and out-of-distribution examples. To avoid using OOD data during training, \citet{2017arXiv171109325L} additionally proposed to use a GAN \citep{Goodfellow:2014:GAN:2969033.2969125} to generate examples near the in-distribution data and force the DNN to produce a uniform distribution at the output for those examples. \citet{hendrycks2019oe} substituted the GAN framework with the Outlier Exposure (OE) technique. More specifically, using a similar loss function, they experimentally showed that by using a real and diverse dataset instead of GAN generated examples, the OOD detection performance of a DNN can be further improved. Recently, \citet{anonymous2020simultaneous} adopted the OE technique and proposed a new loss function that can be used during the training process of a DNN. More specifically, they proposed to substitute the KL divergence metric with the $l_1$ norm of the distance between the output distribution produced by the softmax layer and the uniform distribution. Additionally, they introduced a second regularization term in their loss function which minimizes the Euclidean distance between the training accuracy of a DNN and the average confidence in its predictions on the training set. Since methods like \citet{hendrycks2019oe} and \citet{2017arXiv171109325L} inevitably reduce the confidence of a DNN when making predictions for in-distribution data, the additional regularization term proposed by \citet{anonymous2020simultaneous} made the DNN to better detect in- and out-of-distribution examples at the low softmax probability levels outperforming the previous methods in both image and text classification tasks. Furthermore, they experimentally showed that the addition of this regularization term improved the final test accuracy of the DNN on in-distribution examples. This method is known as Outlier Exposure with Confidence Control (OECC). Recently, \citet{hendrycks2019using} proposed a self-supervised learning approach that performs well on detecting outliers which are close to the in-distribution data but their proposed method does not provide results comparable to \citet{hendrycks2019oe} and \citet{anonymous2020simultaneous} in the general OOD detection task. Note that none of the aforementioned methods requires access to OOD data.

\paragraph{Post-Training Methods for OOD detection.} \citet{Guo:2017:CMN:3305381.3305518} observed that modern neural networks make predictions with much higher confidence compared to their accuracy and they named this phenomenon miscalibration. To mitigate the issue of miscalibration, they proposed the technique of temperature scaling, a variant of the original Platt scaling \citep{PlattProbabilisticOutputs1999}. \citet{2017arXiv170602690L} proposed ODIN as a method to improve the detection performance of a DNN. ODIN used some input pre-processing together with temperature scaling. To tune the parameters of their model, \citet{2017arXiv170602690L} used a sample consisting of both in- and out-of-distribution data. \citet{Lee:2018:SUF:3327757.3327819} proposed another post-training method for OOD detection. More specifically, under the assumption that the pre-trained features of a softmax neural classifier can be fitted well by a class-conditional Gaussian distribution, they defined a confidence score based on Mahalanobis distance to distinguish in- and out-of-distribution samples. Their method used some input pre-processing and a logistic regression classifier to calculate the weight that each layer is going to have in the final Mahalanobis distance-based confidence score. Note that both ODIN and the Mahalanobis Distance-based classifier (MD) require access to a sample of OOD data in order to tune their parameters. Recently, \citet{anonymous2020zeroshot} proposed an alternative post-training OOD detection method that does not require access to OOD samples. More specifically, \citet{anonymous2020zeroshot} proposed the use of higher order Gram matrices to compute pairwise feature correlations and their associated class-conditional bounds on the in-distribution data and experimentally showed that these bounds can be effectively used to distinguish in- and out-of-distribution data during test time. The experimental results showed that this method can outperform the Mahalanobis distance-based classifier method without requiring access to OOD samples in most of the experiments. However, it should be noted that this method does not perform equally well to training methods like OE \citep{hendrycks2019oe} and OECC \citep{anonymous2020simultaneous} whenever it is provided with OOD samples that are close to in-distribution samples as it is the case where CIFAR-10 is used as in-distribution and CIFAR-100 as the OOD set. Since this method calculates Feature Correlations using Gram Matrices for OOD detection, let us call it FCGM method.

\section{OOD Detection Methods}
\subsection{Notation}
Samples used during training are called in-distribution and are generated by a probability distribution $D_{in}$. OE \citep{hendrycks2019oe} and OECC \citep{anonymous2020simultaneous} methods are using an additional dataset during training, called $D_{out}^{OE}$. To tune their hyper-parameters, both methods are using a separate validation dataset consisting of synthetic data, called $D_{out}^{val}$. Finally, OOD samples are considered to be generated by a probability distribution $D_{out}^{test}$. Note that $D_{out}^{OE}$ and $D_{out}^{test}$ are disjoint. Additionally, $D_{out}^{val}$ and $D_{out}^{test}$ are also disjoint.

\subsection{Outlier Exposure with Confidence Control (OECC)}
As also mentioned earlier, OECC method \citep{anonymous2020simultaneous} is a training method for OOD detection. More specifically, based on the technique of Outlier Exposure \citep{hendrycks2019oe}, \citet{anonymous2020simultaneous} proposed to initially train a DNN using the cross-entropy loss function and then fine-tune it using the following loss function:
\begin{equation}
\label{optim3}
\begin{aligned}
\underset{\boldsymbol{\theta}}{\text{minimize}}\hspace{5pt} \mathcal\E_{(x,y)\sim D_{in}}[\mathcal{L}_{CE}(f_{\boldsymbol{\theta}}(x),y)] &+ \lambda_1 \Bigg(A_{tr}-\E_{\rx\sim D_{in}} \Bigg[ \underset{l=1,...,K} \max\Bigg(\frac{e^{z_l}}{\sum_{j=1}^K e^{z_j}}\Bigg) \Bigg]\Bigg)^2 \\&+  \lambda_2 \sum_{x^{(i)} \sim D^{OE}_{out}} \sum_{l=1}^K \Bigg|\frac{1}{K}-\frac{e^{z_l}}{\sum_{j=1}^K e^{z_j}}\Bigg|
\end{aligned}
\end{equation}
where $\lambda_1$ and $\lambda_2$ are hyper-parameters that are tuned using a separate validation dataset $D_{out}^{val}$ and $A_{tr}$ is the training accuracy of the DNN after the training stage using only cross-entropy loss has finished. Note that the first regularization term of (\ref{optim3}) minimizes the Euclidean distance between the training accuracy of the DNN and the average confidence in its predictions on the training set while the second regularization term minimizes the $l_1$ norm of the distance between the output distribution produced by the softmax layer of a DNN and the uniform distribution. \citep{anonymous2020simultaneous} experimentally showed that the loss function descibed by (\ref{optim3}) outperforms the original OE method \citep{hendrycks2019oe} in both image and text classification tasks.   

\subsection{Mahalanobis Distance-based Confidence Score (MD)}
\citet{Lee:2018:SUF:3327757.3327819} proposed a post-training method for OOD detection based on a Mahalanobis distance confidence score. More specifically, they defined the confidence score using the Mahalanobis distance with respect to the closest class-conditional probability distribution. The parameters of this distribution are calculated as empirical class means and tied covariances of the training samples \citep{Lee:2018:SUF:3327757.3327819}. To increase the confidence score of their method, they initially applied input pre-processing by adding a small perturbation at each input example. Subsequently, they extracted all the hidden features of the DNN, calculated their empirical class means and tied covariances and then computed a Mahalanobis distance confidence score for each hidden layer of the DNN. Last, they trained a logistic regression detector using validation samples in order to calculate a weight for each layer's confidence score and finally used those weights to calculate a weighted average of the confidence scores of all layers. Note that MD method requires access to samples from $D_{out}^{test}$ to tune its parameters. 

\subsection{OOD Detection with Feature Correlations using Gram Matrices (FCGM)}
Recently, \citet{anonymous2020zeroshot} proposed a post-training method for OOD detection that does not require access to OOD data for hyper-parameter tuning as MD method \citep{Lee:2018:SUF:3327757.3327819} does. \citet{anonymous2020zeroshot} proposed the use of higher order Gram matrices to compute pairwise feature correlations between the channels of each layer of a DNN. Subsequently, after computing the minimum and maximum values of the correlations for every class $c$ that an example generated by $D_{in}$ is classified, they used those values to calculate the layerwise deviation of each test sample, i.e. the deviation of test sample from the images seen during training with respect to each of the layers. Finally, they calculated the total deviation by taking a normalized sum of the layerwise deviations and using a threshold $\tau$, they classified a sample as OOD if its corresponding total deviation is above the threshold. The experimental results presented in \citet{anonymous2020zeroshot}, showed that FCGM method outperforms MD method in most of the experiments without having access to OOD samples to tune its parameters. However, it should be noted that FCGM, in its current state, does not perform equally well when the samples from $D_{out}^{test}$ are close to $D_{in}$, as it happens for instance in the case where CIFAR-10 is used as $D_{in}$ and CIFAR-100 is used as $D_{out}^{test}$.

\section{Experiments}
For our experiments, we used the publicly available codes of \citet{anonymous2020zeroshot}, \citet{hendrycks2019oe}, \citet{Lee:2018:SUF:3327757.3327819} and \citet{anonymous2020simultaneous}. 

\subsection{Evaluation Metrics}
In our experiments, we adopt the OOD detection evaluation metrics used in \citet{Lee:2018:SUF:3327757.3327819} and \citet{anonymous2020zeroshot}. More specifically, considering an example generated by $D_{in}$ as positive and an example generated by $D_{out}^{test}$ as negative, we define the following evaluation metrics:

\addtolength{\leftskip}{0mm}
  \begin{itemize}[leftmargin=\dimexpr\parindent+0mm+0.5\labelwidth\relax]
  \vspace{-5pt}
  \item True Negative Rate at $N\%$ True Positive Rate ({\it TNRN}):
  This performance metric measures the capability of an OOD detector to detect true negative examples when the true positive rate is set to $95\%$.
  \item Area Under the Receiver Operating Characteristic curve (AUROC): In the out-of-distribution detection task, the ROC curve \citep{Davis:2006:RPR:1143844.1143874} summarizes the performance of an OOD detection method for varying threshold values.
  \item Detection Accuracy (DAcc): As also mentioned in \citet{Lee:2018:SUF:3327757.3327819}, this evaluation metric corresponds to the maximum classification probability over all possible thresholds $\epsilon$:
  \begin{equation*}
  1 - \underset{\epsilon}{\text{min}}\{D_{in}(q(\boldsymbol{x})\leq \epsilon)P(\boldsymbol{x}\text{ is from } D_{in}) + D_{out}(q(\boldsymbol{x}) > \epsilon)P(\boldsymbol{x}\text{ is from } D_{out})\},
  \end{equation*}
  where $q(\boldsymbol{x})$ is a confidence score. Similar to \citet{Lee:2018:SUF:3327757.3327819}, we assume that $P(\boldsymbol{x}\text{ is from } D_{in}) = P(\boldsymbol{x}\text{ is from } D_{out})$. 
\end{itemize}

\subsection{A Combination of OECC and MD Methods for OOD Detection}\label{first_combination}
In these experiments, we demonstrate how the combination of the OECC method \citep{anonymous2020simultaneous}, which is a training method for OOD detection, and the MD method which is post-training method outperforms the results of the original MD method \citep{Lee:2018:SUF:3327757.3327819}. We train ResNet \citep{DBLP:journals/corr/HeZRS15} with 34 layers using CIFAR-10, CIFAR-100 \citep{Krizhevsky09learningmultiple} and SVHN \citep{37648} datasets as $D_{in}$. For the CIFAR experiments, SVHN, TinyImageNet (a sample of 10,000 images drawn from the ImageNet dataset) and LSUN are used as $D_{out}^{test}$. For the SVHN experiments, CIFAR-10, TinyImageNet and LSUN are used as $D_{out}^{test}$. Both TinyImageNet and LSUN images are downsampled to $32\times32$. 

For the results related to the MD method, we train the ResNet model with 34 layers for 200 epochs with batch size 128 by minimizing the cross entropy loss using the SGD algorithm with momentum 0.9. The learning rate starts at 0.1 and is dropped by a factor of 10 at 50\% and 75\% of the training progress, respectively. Subsequently, we compute the Mahalanobis distance-based confidence score using both the input pre-processing and the feature ensemble techniques, where the parameters of the algorithm are tuned using a validation dataset consisting of both in- and out-of-distribution samples similar to \citet{Lee:2018:SUF:3327757.3327819}. For the results related to the combined OECC$+$MD method, we initially trained the ResNet model using only the cross-entropy loss function with exactly the same training details, then we fine-tuned it using the loss function described by (\ref{optim3}) and finally we applied the MD method. During fine-tuning, we used the SGD algorithm with momentum 0.9 and a cosine learning rate \citep{loshchilov-ICLR17SGDR} with an initial value 0.001 using a batch size of 128 for data sampled from $D_{in}$ and a batch size of 256 for data sampled from $D_{out}^{OE}$. In our experiments, the 80 Million Tiny Images dataset \citep{Torralba:2008:MTI:1444381.1444403} was considered as $D_{out}^{OE}$. For CIFAR-10, we fine-tuned the network for 30 epochs, for CIFAR-100 for 20 epochs, while for SVHN the corresponding number of epochs was 5. Both $\lambda_1, \lambda_2$ as well as the parameters of the MD method were tuned using a separate validation dataset consisting of both in- and out-of-distribution samples similar to \citet{Lee:2018:SUF:3327757.3327819}. Note that originally, the OECC method, similar to the OE method \citep{hendrycks2019oe}, does not require access to OOD data for hyper-parameter tuning. However, since the MD method requires access to OOD samples, we tuned all the parameters using the same validation dataset. The results are presented in Table~\ref{Mahalanobis}.  

\begin{table}[t]
\begin{adjustbox}{max width=\textwidth}
\begin{tabular}{cc|cc|cc|cc}
\multicolumn{2}{c}{}&\multicolumn{2}{c}{TNR95$\uparrow$}&\multicolumn{2}{c}{AUROC$\uparrow$}&\multicolumn{2}{c}{DAcc$\uparrow$}\\
\cline{3-8} 
${D}_{in}$&${D}_{out}^{test}$&MD&OECC+MD&MD&OECC+MD&MD&OECC+MD\\
\hline
\multirow{3}{*}{{{CIFAR-10}}}&SVHN&96.4&\textbf{97.3}&99.1&\textbf{99.2}&95.8&\textbf{96.3}\\
&TinyImageNet&97.1&\textbf{98.8}&99.5&\textbf{99.6}&96.3&\textbf{97.3}\\
&LSUN&98.9&\textbf{99.7}&99.7&\textbf{99.8}&97.7&\textbf{98.5}\\
\hline
\multirow{3}{*}{{{CIFAR-100}}}&SVHN&91.9&\textbf{93.0}&98.4&\textbf{98.7}&93.7&\textbf{94.2}\\
&TinyImageNet&90.9&\textbf{92.3}&98.2&\textbf{98.3}&93.3&\textbf{93.9}\\
&LSUN&90.9&\textbf{95.6}&98.2&\textbf{98.6}&93.5&\textbf{95.4}\\
\hline
\multirow{3}{*}{{{SVHN}}}&CIFAR-10&98.4&\textbf{99.9}&99.3&\textbf{99.9}&96.9&\textbf{99.2}\\
&TinyImageNet&99.9&\textbf{100.0}&99.9&\textbf{100.0}&99.1&\textbf{99.9}\\
&LSUN&99.9&\textbf{100.0}&99.9&\textbf{100.0}&99.5&\textbf{100.0}\\
\hline
\end{tabular}
\end{adjustbox}
\caption{\label{Mahalanobis}Comparison using a ResNet-34 architecture between the Mahalanobis distance-based classifier \citep{Lee:2018:SUF:3327757.3327819} versus the combination of OECC method \citep{anonymous2020simultaneous} with the Mahalanobis method. The hyper-parameters are tuned using a validation dataset of in- and out-of-distribution data similar to \citet{Lee:2018:SUF:3327757.3327819}. This table is originally presented in \cite{anonymous2020simultaneous}.}
\end{table}


\subsection{A Combination of OECC and FCGM Methods for OOD Detection}
In these experiments, we show that the combination of the OECC method \citep{anonymous2020simultaneous} together with the FCGM method \citep{anonymous2020zeroshot} outperforms the results presented in \citet{anonymous2020zeroshot} demonstrating again the necessity for a combination between training and post-training methods for OOD detection. The datasets used for these experiments are identical to the ones presented in Section~\ref{first_combination}.
\vspace{-10pt}
\paragraph{ResNet experiments.}For the results related to the FCGM method, we initially trained the ResNet model using exactly the same training details presented in Section~\ref{first_combination} and then we applied the FCGM method where the tuning of the normalizing factor used to calculate the total deviation of a test image is done using a randomly selected validation partition from $D_{in}^{test}$ as described in \citet{anonymous2020zeroshot}. For the combined OECC$+$FCGM method, we initially trained the ResNet model as described above, then we fine-tuned it using the loss function described by (\ref{optim3}) and finally, we applied the FCGM method. During fine-tuning, we used the SGD algorithm with momentum 0.9 and a cosine learning rate \citep{loshchilov-ICLR17SGDR} with an initial value 0.001 using a batch size of 128 for data sampled from $D_{in}$ and a batch size of 256 for data sampled from $D_{out}^{OE}$. In our experiments, the 80 Million Tiny Images dataset \citep{Torralba:2008:MTI:1444381.1444403} was considered as $D_{out}^{OE}$. For CIFAR-10 experiments, we fine-tuned the network for 30 epochs, for CIFAR-100 for 10, while for SVHN the corresponding number of epochs was 5. In contrast with the previous experiment where we combined the OECC method with the MD method, in this experiment, the hyper-parameters $\lambda_1$ and $\lambda_2$ of (\ref{optim3}) were tuned using a separate validation dataset as described in Appendix~\ref{val_image}. Note that $D_{out}^{val}$ and $D_{out}^{test}$ are disjoint. Therefore, for these experiments, no access to $D_{out}^{test}$ was assumed. The results of the experiments are shown in Table~\ref{FCorr}.

\begin{table}[t]
\begin{adjustbox}{max width=\textwidth}
\begin{tabular}{cc|cc|cc|cc}
\multicolumn{2}{c}{}&\multicolumn{2}{c}{TNR95$\uparrow$}&\multicolumn{2}{c}{AUROC$\uparrow$}&\multicolumn{2}{c}{DAcc$\uparrow$}\\
\cline{3-8} 
${D}_{in}$&${D}_{out}^{test}$&FCGM&OECC+FCGM&FCGM&OECC+FCGM&FCGM&OECC+FCGM\\
\hline
\multirow{3}{*}{{{CIFAR-10}}}&SVHN&97.6&\textbf{99.2}&99.4&\textbf{99.7}&96.7&\textbf{98.0}\\
&TinyImageNet&98.7&\textbf{99.6}&99.6&\textbf{99.8}&97.8&\textbf{98.3}\\
&LSUN&99.6&\textbf{99.9}&99.8&\textbf{99.9}&98.6&\textbf{99.0}\\
\hline
\multirow{3}{*}{{{CIFAR-100}}}&SVHN&81.4&\textbf{87.2}&96.2&\textbf{97.1}&89.8&\textbf{91.9}\\
&TinyImageNet&95.1&\textbf{95.8}&\textbf{99.0}&98.8&95.1&\textbf{95.5}\\
&LSUN&97.0&\textbf{98.2}&99.3&99.3&96.2&\textbf{96.8}\\
\hline
\multirow{3}{*}{{{SVHN}}}&CIFAR-10&85.7&\textbf{98.3}&97.3&\textbf{99.3}&91.9&\textbf{96.9}\\
&TinyImageNet&99.3&\textbf{100.0}&99.7&\textbf{100.0}&97.9&\textbf{99.5}\\
&LSUN&99.4&\textbf{100.0}&99.8&\textbf{100.0}&98.5&\textbf{99.8}\\
\hline
\end{tabular}
\end{adjustbox}
\caption{\label{FCorr}Comparison using a ResNet-34 architecture between the zero-shot OOD detection with feature correlations method proposed by \citep{anonymous2020zeroshot} versus the combination of OECC method \citep{anonymous2020simultaneous} and the FCGM method. The tuning of the hyperparameters $\lambda_1$ and $\lambda_2$ of \citet{anonymous2020simultaneous} is done using a separate validation dataset $D_{val}^{out}$ presented in Appendix~\ref{val_image}. Note that $D_{out}^{val}$ and $D_{out}^{test}$ are disjoint.}
\end{table}
\vspace{-10pt}
\paragraph{DenseNet experiments.} For the results related to the FCGM method, we used the pre-trained DenseNet \citep{huang2017densely} model provided by \citet{2017arXiv170602690L}. The network has depth $L=100$, growth rate $m=12$ and dropout rate 0. It has been trained using the stochastic gradient descent algorithm with Nesterov momentum \citep{Duchi:2011:ASM:1953048.2021068, kingma2014adam} for 300 epochs with batch size 64 and momentum 0.9. The learning rate started at 0.1 and was dropped by a factor of 10 at $50\%$ and $75\%$ of the training progress, respectively. Subsequently, we applied the FCGM method \citep{anonymous2020zeroshot} where the tuning of the normalizing factor used to calculate the total deviation of a test image was done using a randomly selected validation partition from $D_{in}^{test}$ as described in \citet{anonymous2020zeroshot}. For the combined OECC+FCGM method, we fine-tuned the pre-trained DenseNet network model provided by \citep{2017arXiv170602690L} using the OECC loss function \citep{anonymous2020simultaneous} described by (\ref{optim3}) and then we applied the FCGM method. During fine-tuning, we used the SGD algorithm with momentum 0.9 and a cosine learning rate \citep{loshchilov-ICLR17SGDR} with an initial value 0.001 for CIFAR-10 and SVHN experiments and 0.01 for the CIFAR-100 experiments using a batch size of 128 for data sampled from $D_{in}$ and a batch size of 256 for data sampled from $D_{out}^{OE}$. In our experiments, the 80 Million Tiny Images dataset \citep{Torralba:2008:MTI:1444381.1444403} was considered as $D_{out}^{OE}$. The DenseNet model was fine-tuned for 15 epochs for the CIFAR-10 experiments, for 10 epochs for the CIFAR-100 experiments, while for SVHN the corresponding number of epochs was 5. The hyperparameters $\lambda_1$ and $\lambda_2$ of the OECC method were tuned using a separate validation dataset $D_{out}^{val}$ as described in Appendix~\ref{val_image}. Note that $D_{out}^{val}$ and $D_{out}^{test}$ are disjoint. The experimental results are presented in Table~\ref{FCorr_Densenet}. 

\begin{table}[h]
\begin{adjustbox}{max width=\textwidth}
\begin{tabular}{cc|cc|cc|cc}
\multicolumn{2}{c}{}&\multicolumn{2}{c}{TNR95$\uparrow$}&\multicolumn{2}{c}{AUROC$\uparrow$}&\multicolumn{2}{c}{DAcc$\uparrow$}\\
\cline{3-8} 
${D}_{in}$&${D}_{out}^{test}$&FCGM&OECC+FCGM&FCGM&OECC+FCGM&FCGM&OECC+FCGM\\
\hline
\multirow{3}{*}{{{CIFAR-10}}}&SVHN&96.0&\textbf{98.5}&99.1&\textbf{99.6}&95.8&\textbf{97.4}\\
&TinyImageNet&98.8&\textbf{99.3}&99.7&\textbf{99.8}&97.9&\textbf{98.3}\\
&LSUN&99.5&\textbf{99.8}&99.9&99.9&97.9&\textbf{99.0}\\
\hline
\multirow{3}{*}{{{CIFAR-100}}}&SVHN&\textbf{89.4}&88.9&\textbf{97.4}&97.0&\textbf{92.4}&92.1\\
&TinyImageNet&95.8&\textbf{96.2}&99.0&99.0&95.6&\textbf{95.7}\\
&LSUN&97.3&\textbf{98.1}&\textbf{99.4}&99.3&96.4&\textbf{97.0}\\
\hline
\multirow{3}{*}{{{SVHN}}}&CIFAR-10&80.2&\textbf{98.5}&95.5&\textbf{99.6}&89.0&\textbf{97.5}\\
&TinyImageNet&99.1&\textbf{99.9}&99.7&\textbf{100.0}&97.9&\textbf{99.7}\\
&LSUN&99.5&\textbf{100.0}&99.8&\textbf{100.0}&98.5&\textbf{99.9}\\
\hline
\end{tabular}
\end{adjustbox}
\caption{\label{FCorr_Densenet}Comparison using a DenseNet-100 architecture between the zero-shot OOD detection with feature correlations method proposed by \citep{anonymous2020zeroshot} versus the combination of OECC method \citep{anonymous2020simultaneous} and the FCGM method. The tuning of the hyperparameters $\lambda_1$ and $\lambda_2$ of \citet{anonymous2020simultaneous} is done using a separate validation dataset $D_{val}^{out}$ presented in Appendix~\ref{val_image}. Note that $D_{out}^{val}$ and $D_{out}^{test}$ are disjoint.}
\end{table}

\paragraph{Discussion.}The results presented in Table~\ref{Mahalanobis}, Table~\ref{FCorr} and Table~\ref{FCorr_Densenet} demonstrate the superior performance that can be achieved when combining training and post-training methods for OOD detection. More specifically, the MD method \citep{Lee:2018:SUF:3327757.3327819} extracts the features from all layers of a pre-trained softmax neural classifier and then calculates the Mahalanobis distance-based confidence score. The FCGM method \citep{anonymous2020zeroshot} also extracts the features from a pre-trained softmax neural classifier and then computes higher order Gram matrices to subsequently calculate pairwise feature correlations between the channels of each layer of a DNN. As also mentioned earlier, both of these methods are post-training methods for OOD detection. On the other hand, OECC method \citep{anonymous2020simultaneous} which belongs to the category of training methods for OOD detection, trains a DNN to learn a feature representation at the output of a DNN such that it makes it capable to better distinguish in- and out-of-distribution examples compared to the baseline method \citep{hendrycks17baseline} and the OE method \citep{hendrycks2019oe}. Therefore, by feeding a post-training method like \citet{Lee:2018:SUF:3327757.3327819} and \citet{anonymous2020zeroshot} with better feature presentations, it is expected that one can achieve superior results in the OOD detection task as it is also validated by the experimental results in Table~\ref{Mahalanobis}, Table~\ref{FCorr} and Table~\ref{FCorr_Densenet}.

\section{Conclusion}
In this paper, we reviewed some of the most recent seminal algorithms in OOD detection and we mentioned the necessity for a combination between training and post-training methods for OOD detection. We experimentally showed that by combining the OECC method \citep{anonymous2020simultaneous} with the Mahalanobis distance-based classifier method \citep{Lee:2018:SUF:3327757.3327819}, one can outperform the original results of the Mahalanobis method. Additionally, we showed that the combination of the OECC method with the recent zero-shot OOD detector with feature correlations method \citep{anonymous2020zeroshot} can outperform the original results of the latter. We hope that these experimental findings will push the future research efforts to look for a combination of training and post-training methods for OOD detection.


\subsubsection*{Acknowledgments}
We thank Google for donating Google Cloud Platform research credits used in this research.

\bibliography{iclr2020_conference}
\bibliographystyle{iclr2020_conference}

\clearpage

\appendix
\section{Validation Data $D_{out}^{val}$}\label{val_image}
The synthetic datasets used for validation purposes in our experiments were initially proposed by \citet{hendrycks2019oe}. 

\textbf{Uniform Noise:} A synthetic image dataset where each pixel is sampled from $\mathcal{U}[0,1]$ or $\mathcal{U}[-1,1]$ depending on the input space of the classifier.
\vspace{-5pt}

\textbf{Arithmetic Mean: }A synthetic image dataset created by randomly sampling a pair of in-distribution images and subsequently taking their pixelwise arithmetic mean.
\vspace{-5pt}

\textbf{Geometric Mean: }A synthetic image dataset created by randomly sampling a pair of in-distribution images and subsequently taking their pixelwise geometric mean.
\vspace{-5pt}

\textbf{Jigsaw: }A synthetic image dataset created by partitioning an image sampled from $D_{in}$ into 16 equally sized patches and by subsequently permuting those patches.
\vspace{-5pt}

\textbf{Speckle Noised: }A synthetic image dataset created by applying speckle noise to images sampled from $D_{in}$.
\vspace{-5pt}

\textbf{Inverted Images: }A synthetic image dataset created by shifting and reordering the color channels of images sampled from $D_{in}$.
\vspace{-5pt}

\textbf{RGB Ghosted: }A synthetic image dataset created by inverting the color channels of images sampled from $D_{in}$.

\clearpage

\end{document}